# Robot Voice A Voice Controlled Robot using Arduino

**Vineeth Teeda, K.Sujatha, Rakesh Mutukuru**

*Abstract—Robotic assistants reduces the manual efforts being put by humans in their day-to-day tasks. In this paper, we develop a voice-controlled personal assistant robot. The human voice commands are taken by the robot by it's own inbuilt microphone. This robot not only takes the commands and execute them, but also gives an acknowledgement through speech output. This robot can perform different movements, turns, wakeup/shutdown operations, relocate an object from one place to another and can also develop a conversation with human. The voice commands are processed in real-time, using an offline server. The speech signal commands are directly communicated to the server using a USB cable. The personal assistant robot is developed on a micro-controller based platform. Performance evaluation is carried out with encouraging results of the initial experiments. Possible improvements are also discussed towards potential applications in home, hospitals, car systems and industries.*

## I. INTRODUCTION

Speech signals are the most important means of communication in human beings. Almost every conversation to interact is done by means of voice signals. Sounds and various speech signals can be converted into electrical form using a microphone. Voice recognition is a technology which is used to convert the speech signals into a computer text format. This voice recognition technology can be used to control and generate speech acknowledgement using some external server.

Robotvoice has the ability to understand thousands of voice commands and perform required action. The voice recognition is a bit difficult task because each person has his own accent. For that, Robot voice uses Bit Voicer Server which supports 17 languages from 26 countries and regions. These robotic assistants can be used for shaping, manufacturing and tooling purposes in various sectors such as manufacturing, defence etc. In hospitals, these robotic assistant can be used for the purpose of performing surgeries and operations with high precision. In this paper, we develop an assistant robot that can be operated using speech commands [1].

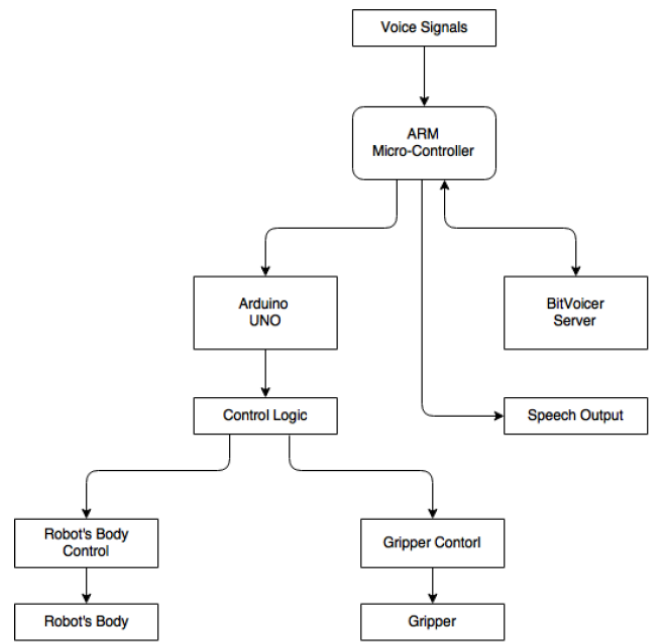

**Fig. 1: Block Diagram of Robot**

Robotvoice has a gripper which is used to pickup and drop the objects. A PC-interfaced low cost robotic arm was developed that could be integrated with a robotic arm, used for light weight lifting applications [2]. Another robotic arm was developed which had special applications for the physically challenged people [3]. A robotic arm was designed in such a way that it could be controlled using human brain [4]. A robotic arm was designed which could drill boreholes and in congested urban areas with effectiveness [5].

The studies have shown that speech is the most rapid form of communication with many fields of applications

- In-car systems.
- Health Care.
- Military.
- Education.
- Intelligent buildings.

Existing system is all about only commanding the robot which understands the commands and perform actions. We want to take this a step forward and add a speech synthesiser which generates speech which is then amplified and given to a speaker.

This paper is organised as follows. In Section II, the construction. A brief discussion on hardware that is used is made in Section III. Section IV discusses the different softwares that are used in Robotvoice. Section V discusses the coding and algorithms.

Section VI discusses about the applications of this robot and section VII discusses about the summary and conclusion of this project.

## II. CONSTRUCTION DETAILS OF ROBOT VOICE

The movement of the robots body is controlled using voice commands i.e, (i) robot's body, (ii) gripper depend on the voice command given. The voice signal is captured using inbuilt microphone and is transmitted over a usb cable to an external offline server, where it is converted into speech format and necessary commands and speech is synthesised. The schematic block diagram is shown in Fig. 1. The hardware platform mainly consists of a gripper which is used to pickup and drop objects, a chassis where micro-controllers and DC motors, motor drivers are fixed. Robot voice has to chassis lower chassis consists of Arduino Uno micro-controller which is used for robot movement operation, DC motors, motor drivers, a speaker, and DC batteries.

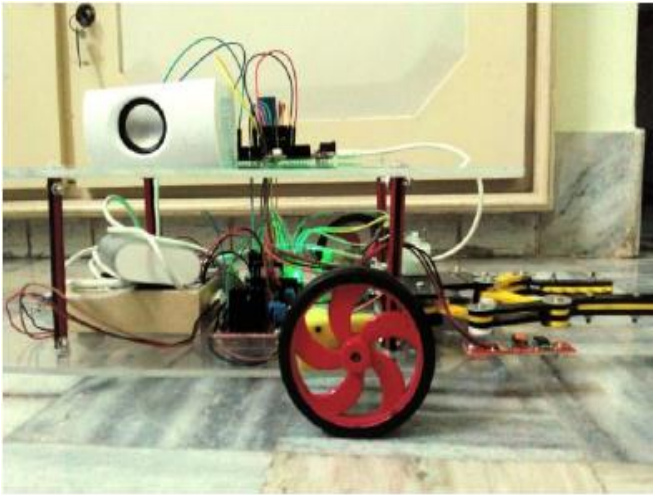

Fig. 2(a): Side view of Robot voice

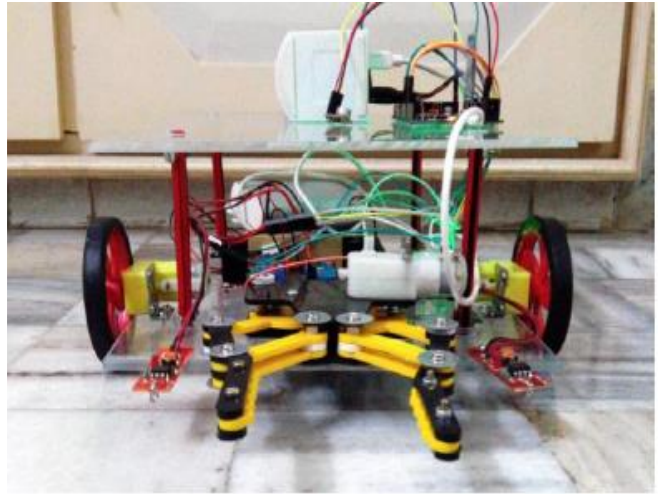

Fig. 2(b): Front view of Robot voice

The upper chassis has an Arduino Due, a microphone and power supplies. Two DC motors are used to control the movement of the robot's body. Another two DC motors are used to control the movement of the robotic arm and the robotic hands. Motor driver IC is used to control the movement of DC motor. A single motor driver (L293D) IC can control two DC motors.

A general USB cable is connected to Arduino Due which interfaces with BitVoicer Server to transfer the speech signals. Depending upon the commands given the robot performs actions accordingly. Movement of the robot's bodyand gripper are independent of each other's movements. Side view of the robot with the robotic hands open is shown in Fig. 2(a), front view of Robotvoice is shown in Fig. 2(b). As shown in the above figure the upper deck of the robot has a Arduino Due which is used for taking the voice commands, synthesising speech, and also forward the commands to the Arduino Uno. Forwarding commands from one micro-controller to another is done by using a pin to pin connection. If a particular pin is made high then that value is forwarded to Arduino Uno. Then the code in Uno checks for the active pin and executes the particular function. For example, if pin 3 is high, the code in Uno is written in such a way that when pin3 is active the robot checks with the IR sensors and then moves forward. Likewise for all the movement operations, each pin is assigned and the robot operation is properly maintained. Power supply to Robotvoice is given by three 9V batteries and the amplifier and speaker is powered by a 5V power bank.

## III. HARDWARE USED IN ROBOT VOICE

Various hardware that are used in this project are,
- Arduino Due.
- Arduino Uno.
- Spark Fun Electret Microphone Breakout.
- Audio Amplifier(HT6818).
- DC motors.
- L293D motor drivers.
- Gripper.
- IR Sensors.
- Speaker

Now, let us learn about each of the hardware modules used briefly.

### A. Arduino Due:

The Arduino Due is a micro-controller board based on the Atmel SAM3X8E ARM Cortex-M3 CPU. It is the first Arduino board based on a 32-bit ARM core micro-controller. It has 54 digital input/output pins (of which 12 can be used as PWM outputs), 12 analog inputs, 4 UARTs (hardware serial ports), a 84 MHz clock, an USB OTG capable connection, 2 DAC (digital to analog), 2 TWI, a power jack, an SPI header, a JTAG header, a reset button and an erase button. It has an inbuilt digital to analog converter which is used in speech synthesis.

The board contains everything needed to support the micro-controller; simply connect it to a computer with a micro-USB cable or power it with a AC-to-DC adapter or battery to get started. The Due is compatible with all Arduino shields that work at 3.3V and are compliant with the 1.0 Arduino pinout.

The SAM3X has 512 KB (2 blocks of 256 KB) of flash memory for storing code. The boot loader is pre burned in factory from Atmel and is stored in a dedicated ROM memory. The available SRAM is 96 KB in two contiguous bank of 64 KB and 32 KB. All the available memory (Flash, RAM and ROM) can be accessed directly as a flat addressing space.It is possible to erase the Flash memory of the SAM3X with the onboard erase button. This will remove the currently loaded sketch from the MCU. To erase, press and hold the Erase button for a few seconds while the board is powered.

### B. Arduino Uno:

The Uno is a micro-controller board based on the ATmega328P. It has 14 digital input/output pins (of which 6 can be used as PWM outputs), 6 analog inputs, a 16 MHz quartz crystal, a USB connection, a power jack, an ICSP header and a reset button. It contains everything needed to support the micro-controller; simply connect it to a computer with a USB cable or power it with a AC-to-DC adapter or battery to get started.The ATmega328 has 32 KB (with 0.5 KB occupied by the boot loader). It also has 2 KB of SRAM and 1 KB of EEPROM.

The Uno has a number of facilities for communicating with a computer, another Uno board, or other micro-controllers. The ATmega328 provides UART TTL (5V) serial communication, which is available on digital pins 0 (RX) and 1 (TX). An ATmega16U2 on the board channels this serial communication over USB and appears as a virtual com port to software on the computer. The 16U2 firmware uses the standard USB COM drivers, and no external driver is needed. However, on Windows, a .inf file is required. The Arduino Software (IDE) includes a serial monitor which allows simple textual data to be sent to and from the board. The RX and TX LEDs on the board will flash when data is being transmitted via the USB-to-serial chip and USB connection to the computer (but not for serial communication on pins 0 and 1).

### C. SparkFun Electret Microphone Breakout:

This small breakout board couples a small electret microphone with a 100x opamp to amplify the sounds of voice, door knocks, etc loud enough to be picked up by a micro controller's Analog to Digital converter. Unit comes fully assembled as shown. Works from 2.7V up to 5.5V. The main specification of opamp is it amplifies bot AC and DC signals. The OPA344 and OPA345 series rail-to-rail CMOS operational amplifiers are designed for precision, low-power, miniature applications. The OPA344 is unity gain stable, while the OPA345 is optimised for gains greater than or equal to five, and has a gain-bandwidth product of 3MHz.

### D. Audio Amplifier (HT6818):

A class-D amplifier or switching amplifier is an electronic amplifier in which the amplifying devices (transistors, usually MOSFETs) operate as electronic switches, instead of as linear gain devices as in other amplifiers. The signal to be amplified is a train of constant amplitude pulses, so the active devices switch rapidly back and forth between a fully conductive and nonconductive state. The analog signal to be amplified is converted to a series of pulses by pulse width modulation, pulse density modulation or other method before being applied to the amplifier. It is a class D stereo amplifier which can drive a speaker upto 8 ohms impedance and 5 Watts. A class-D amplifier or switching amplifier is an electronic amplifier in which the amplifying devices (transistors, usually MOSFETs) operate as electronic switches, instead of as linear gain devices as in other amplifiers.

### E. DC Motors:

A DC motor is an electric motor that runs on direct current (DC) electricity. DC motors were used to run machinery, often eliminating the need for a local steam engine or internal combustion engine. DC motors can operate directly from rechargeable batteries, providing the motive power for the first electric vehicles. Today DC motors are still found in applications as small as toys and disk drives, or in large sizes to operate steel rolling mills and paper machines. Modern DC motors are nearly always operated in conjunction with power electronic devices. 300 RPM Side Shaft Heavy Duty DC Gear Motor is suitable for large robots / automation systems. It has sturdy construction with gear box built to handle stall torque produced by the motor. Drive shaft is supported from both sides with metal bushes. Motor runs smoothly from 4V to 12V and gives 300 RPM at 12V. Motor has 8mm diameter, 17.5mm length drive shaft with D shape for excellent coupling.

### F. L293D motor drivers:

Thisdevice is a monolithic integrated high voltage, high current four channel driver designed to accept standard DTL or TTL logic levels and drive inductive loads (such as relays solenoids, DC and stepping motors) and switching power transistors. Each pair of channels is equipped with an enable input. A separate supply input is provided for the logic, allowing operation at a lower voltage and internal clamp diodes are included.

This device is suitable for use in switching applications at frequencies up to 5 kHz. The Motor Shield is able to drive 2 servo motors, and has 8 half-bridge outputs for 2 stepper motors or 4 full H-bridge motor outputs or 8 half-bridge drivers, or a combination.The servo motors use the +5V of the Arduino board. The voltage regulator on the Arduino board could get hot. To avoid this, the newer Motor Shields have connection points for a separate +5V for the servo motors.

### G. Gripper:

The Gripper module is state of robotic arm which can be used in various 'pick and place' kind of robots. It works on DC Motor (5 to 12V DC).

# Robot Voice A Voice Controlled Robot using Arduino

Change in rotation direction of the DC Motor, generates Jaw Open & Close Action. The DC motor can be easily be controlled with the help of DPDT Switch (manual mode) or with the help of any micro-controller along with L293D Motor Driver module.It can be used in various 'pick and place' kind of robots. It works on DC Motor (9 to 12V DC). Change in rotation direction of the DC Motor, generates Jaw Open & Close Action. The DC motor can be easily be controlled with the help of DPDT Switch (manual mode) or with the help of any micro-controller along with L293D Motor Driver module. Gives an extra functionality to your robots by adding a fully functional Robot gripper to them.

*H. IR Sensor:*

An infrared sensor is an electronic device, that emits in order to sense some aspects of the surroundings. An IR sensor can measure the heat of an object as well as detects the motion.These types of sensors measures only infrared radiation, rather than emitting it that is called as a passive IR sensor.This sensor is analogous to human's visionary senses, which can be used to detect obstacles and it is one of the common applications in real time.This circuit comprises of the following components. An object moving nearby actually reflects the infrared rays emitted by the infrared LED. The infrared receiver has sensitivity angle (lobe) of 0-60 degrees, hence when the reflected IR ray is sensed, the mono in the receiver part is triggered.

- LM358 IC 2 IR transmitter and receiver pair.
- Resistors of the range of kilo ohms.
- Variable resistors.
- LED (Light Emitting Diode).

*I. Speaker:*

Loudspeaker or speaker, device used to convert electrical energy into sound. It consists essentially of a thin flexible sheet called a diaphragm that is made to vibrate by an electric signal from an amplifier. The vibrations create sound waves in the air around the speaker. In a dynamic speaker, the most common kind, the diaphragm is cone-shaped and is attached to a coil of wire suspended in a magnetic field produced by a permanent magnet. A signal current in the suspended coil, called a voice coil, creates a magnetic field that interacts with the already-existing field, causing the coil and the diaphragm attached to it to vibrate. To provide a faithful reproduction of music or speech, a loudspeaker must be able to reproduce a wide range of audio frequencies (i.e., 20 Hz to 20 kHz).

## IV. SOFTWARE USED IN ROBOT VOICE

*A. Arduino IDE:*

The Arduino Integrated Development Environment - or Arduino Software (IDE) - contains a text editor for writing code, a message area, a text console, a toolbar with buttons for common functions and a series of menus. It connects to the Arduino and Genuino hardware to upload programs and communicate with them. Programs written using Arduino Software (IDE) are called sketches. These sketches are written in the text editor and are saved with the file extension .ino. The editor has features for cutting/pasting and for searching/replacing text. The message area gives feedback while saving and exporting and also displays errors. The console displays text output by the Arduino Software (IDE), including complete error messages and other information. The bottom righthand corner of the window displays the configured board and serial port. The toolbar buttons allow you to verify and upload programs, create, open, and save sketches, and open the serial monitor.

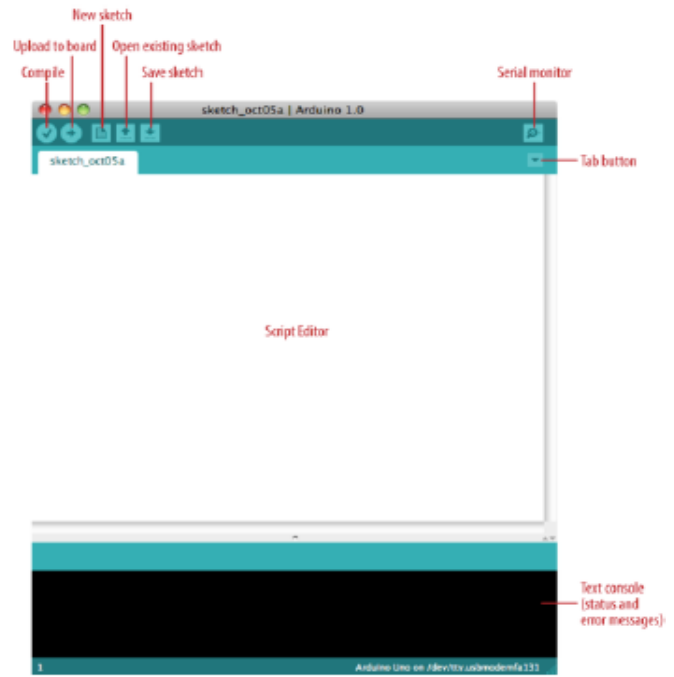

**Fig. 4(a): Arduino IDE main window**

The open-source Arduino Software (IDE) makes it easy to write code and upload it to the board. It runs on Windows, Mac OS X, and linux.The Arduino project provides the Arduino integrated development environment (IDE), which is a cross-platform application written in Java. It originated from the IDE for the Processing programming language project and the Wiring project. It is designed to introduce programming to artists and other newcomers unfamiliar with software development. It includes a code editor with features such as syntax highlighting, brace matching, and automatic indentation, and provides simple one-click mechanism for compiling and loading programs to an Arduino board. A program written with the IDE for Arduino is called a "sketch" [6]. The main window of Arduino IDE is shown in Fig. 4(a). The Arduino Software (IDE) uses the concept of a sketchbook: a standard place to store your programs (or sketches). The sketches in your sketchbook can be opened from the File > Sketchbook menu or from the Open button on the toolbar. The first time you run the Arduino software, it will automatically create a directory for your sketchbook.

You can view or change the location of the sketchbook location from with the Preferences dialog.Libraries provide extra functionality for use in sketches, e.g. working with hardware or manipulating data. To use a library in a sketch, select it from the Sketch > Import Librarymenu.

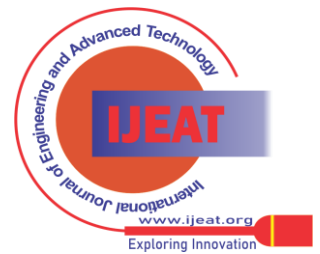

This will insert one or more #include statements at the top of the sketch and compile the library with your sketch. Because libraries are uploaded to the board with your sketch, they increase the amount of space it takes up. There is a list of libraries in the reference. Some libraries are included with the Arduino software. Others can be downloaded from a variety of sources or through the Library Manager. Starting with version 1.0.5 of the IDE, you do can import a library from a zip file and use it in an open sketch.

*B. Bit Voicer Server:*

BitVoicer Server is a speech recognition and synthesis server for speech automation. It was developed to enable simple devices, with low processing power, to become voice-operated. In general, micro-controllers do not have enough memory and processing power to perform advanced speech recognition and synthesis. BitVoicer Server eliminates the effects of these limitations by performing the hard work so the micro-controller can allocate most of its resources to its core functionality. Fig. 4(b) shows the main window of the BotVoicer Server.

Input Devices are those capable of capturing, digitising and sending audio streams to the server. When BitVoicer Server identifies an Input Device, it assigns one exclusive Speech Recognition Engine (SRE) to that device. SREs constantly analyse all audio streams sent to the server and when a predefined sentence is identified, BitVoicer Server performs the actions specified by the user. These actions are called commands and can be used to start other applications, synthesise speech, play audio files or send data (commands) to Output and Mixed Devices. The user can define one or more commands for each sentence.

of the micro-controller that is connected to the BitVoicer Server. It also has the options of volume of the synthesised speech, minimum audio level to be recognised from the microphone, type of the micro-controller which includes input, output or mixed. In this case we are using a mixed type of micro-controller. Serial number column is used only when we are using multiple micro-controllers. This is shown in Fig. 4(d). Next is binary data where we add the data that should be sent to the micro-controller.

The BitVoicer Server supports the following languages,

- Catalan (Catalonia).
- Chinese (China, Honk Kong and Taiwan).
- Danish (Denmark).
- Dutch (Netherlands).
- English (Australia, Canada, India, UK and US).
- Finnish (Finland).
- French (Canada and France).
- German (Germany).
- Italian (Italy).
- Japanese (Japan).
- Korean (Korea).
- Norwegian, Bokmål (Norway).
- Polish (Poland).
- Portuguese (Brazil and Portugal).
- Russian (Russia).
- Spanish (Mexico and Spain).
- Swedish (Sweden).

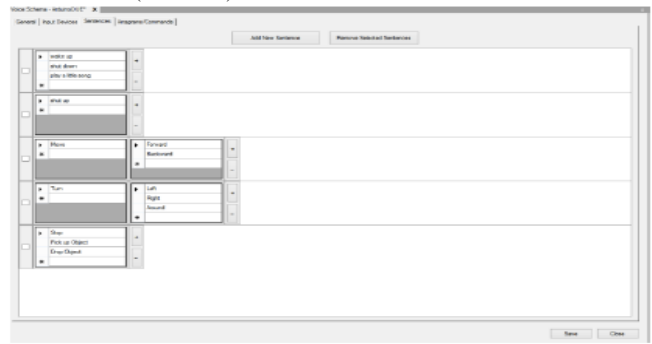

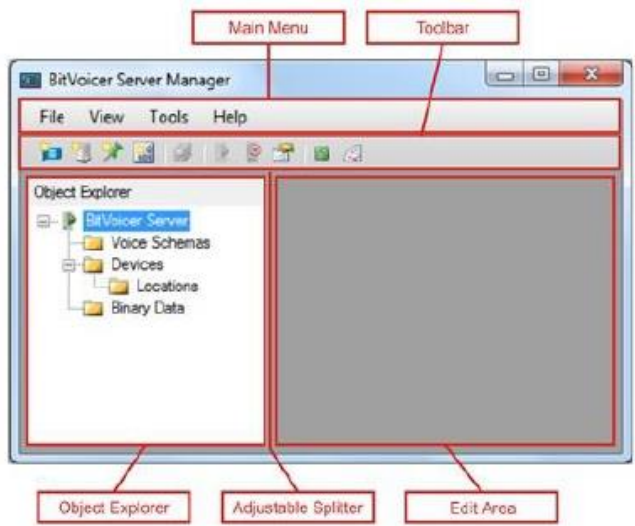

**Fig. 4(b): BitVoicer Server main window**

The user can also define the order in which the commands will be executed, the time interval between them, and which Output or Mixed Devices are the targets of the commands. That is, with one single Input Device license you can control multiple Output Devices. Lastly, there are the Mixed Devices that are capable of acting as Input and Output Devices [7]. The BitVoicer Server consists of Voice Schemas where we add all the sentences that are received from microphone and also the speech that should be synthesised. This is shown in Fig. 4(c). Next we can see Devices in the main window. This window shows the details

**Fig. 4(c): Voice Schemas window**

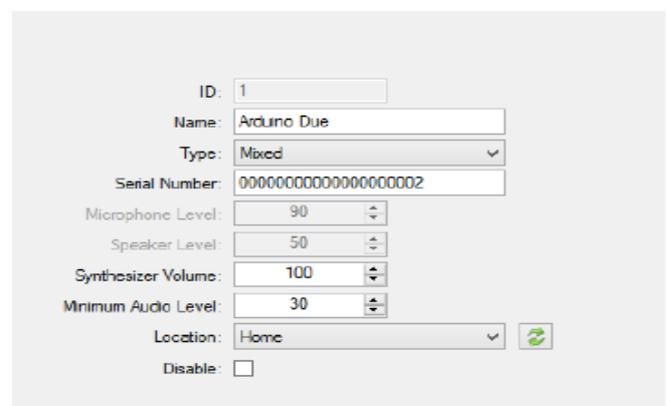

**Fig. 4(d): Devices Window**

# Robot Voice A Voice Controlled Robot using Arduino

It has the following features-
- Multi-device: BitVoicer Server has a new manager for speech recognition and synthesis engines which is able to serve multiple devices simultaneously. Each device can function as a point of capture or reproduction and have its exclusive speech recognition or synthesis engine.
- Enhanced protocol: The BitVoicer Server Protocol (BVSP) brings significant improvements over the communication protocol used in the previous BitVoicer versions. Now client devices are able to retrieve server status information and exchange data with other applications running on the server through the same communication channel used to transmit audio.
- Windows service: Besides consuming few hardware resources, BitVoicer Server is executed as a Windows service. This enables BitVoicer Server to become a transparent resource and cause little or no interference on the user interface.
- Integration: Application developers can retrieve the results of recognition operations from the server and exchange data with client devices through Windows Communication Foundation (WCF) services. For .NET development, there is also a .NET integration library available.
- It does not require speech recognition enginetraining.
- It does not require pre-recording of sentences.
- Unlimited number of sentences, commands anddevices.

## V. ALGORITHMS AND CODING

As we are using two micro-controllers, one for communication between the Robotvoice and server and another for the movement of the robot. Arduino Due is used for communication and Arduino Uno for movement operation. The algorithm for Arduino Due i.e, to establish and communicate with server is given in Fig. 5(a). The basic operation here is nothing but if server is available or is free then, robot starts collecting the speech signals through microphone and sends it to the server. Whereas when the server is busy it waits for the server to send any data or speech signals that are synthesised in the server. The received data is generally in binary format which specifies which pin to be high or low. If the data is a speech signal then Arduino Due converts the electrical signal into analog form and forwards it to amplifier, which is later on amplified and sent to speaker. Here the connection between the two micro-controllers is a port to port connection. Each pin of Arduino Due is connected to a specific pin of Arduino Uno. If a particular port is made high in Due, the same is carried to Uno which performs a specific function which is allocated to it.

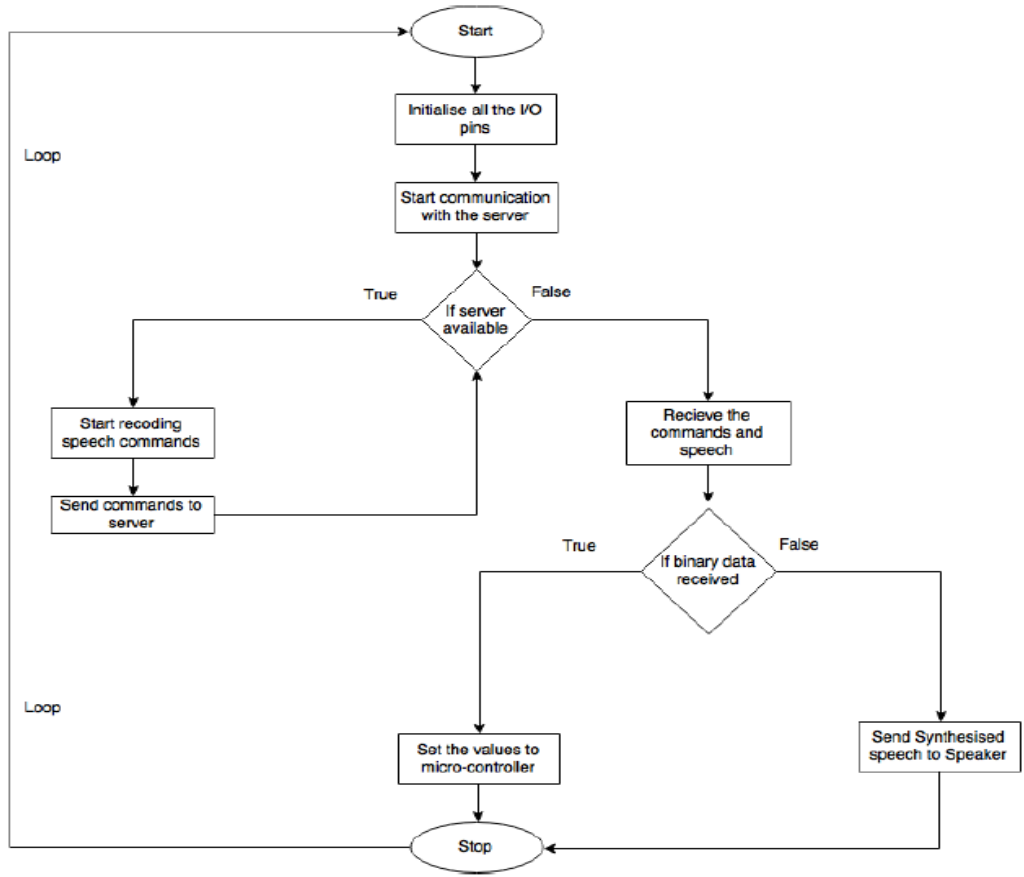

**Fig. 5(a): Algorithm of Arduino Due code**

Another micro-controller that is used in this project is Arduino uno which takes care of all the movement operations. We design the code in such a way that each movement operation is done when a specific pin is active and that is provided by Arduino Due.

## VI. RESULTS

The following are the pictures taken while the robot is in action. The robot has two sections for the voice recognition and synthesis. The upper part as you can see consists of a small microphone, Arduino Due and speaker. The lower deck consists of Arduino Uno, motor drivers, dc motors, gripper, IR sensors and power supplies. Fig. 6(a) shows the top view of Robot Voice, when it is in action.

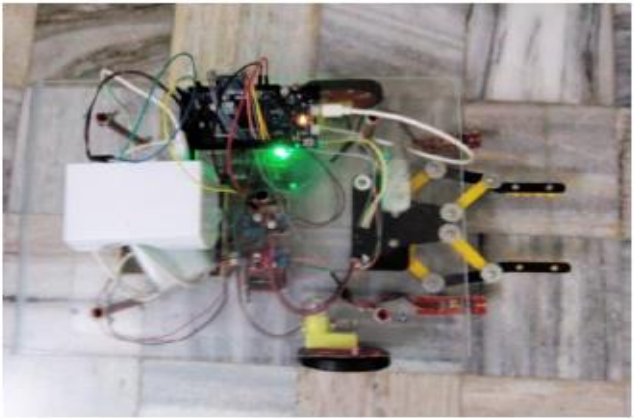

**Fig. 6 (a): Top view of Robot Voice**

Here the robot is connected to an external server where all the work is done. Here that connection is established using a USB cable. Fig. 6(b) shows the robot holding an object by taking commands from the user.

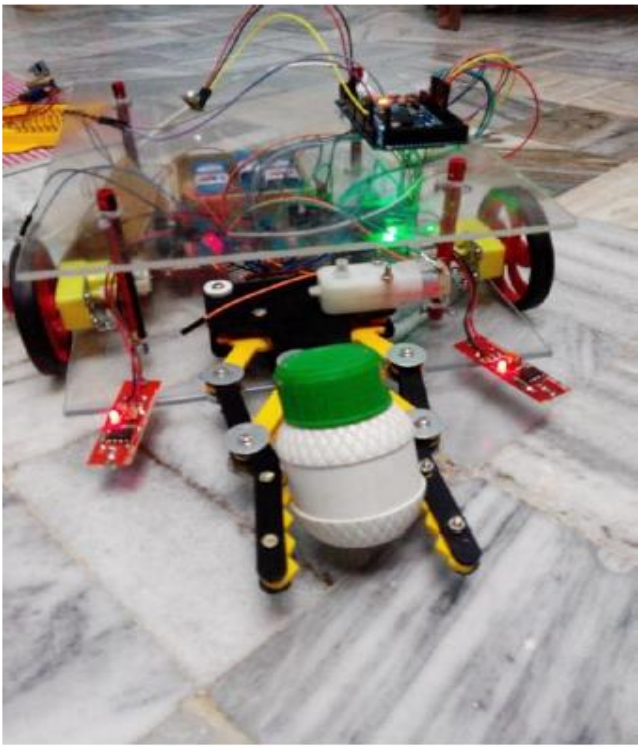

**Fig. 6(b): Robot Voice holding an object using gripper**

## VII. APPLICATIONS OF THE ROBOTVOICE

The assistant robot can be used for various purposes as listed below:

### A. In chemical industries:

In chemical industry, people cannot handle the chemicals which might be having high temperature. Thus, industrial robot is a vital development towards improving the safety standards in industries [9]. In such hazardous situations, the assistant robot's can be used to hold the chemicals and carry them from one place to another without human interference. Also, there might be places in the industries where humans cannot go and work, in all such cases this robot can be controlled by the voice commands and can be directed to go and work in that place. It can also be used to carry small objects in the industry within a certain distance to reduce the time and the manual labour. Robotic assistant can also be used in manufacturing sector for different re-positioning operations.

### B. In homes and for daily needs:

People may need assistance to reduce their manual effort, which may be mostly needed in the case of physically handicapped people or the old-aged people. Robotic assistant can be used by physically challenged people or the old-aged people as it helps them to place an object from one place to another which would be difficult for them in general. These assistant robot's can move around quickly and also can be controlled easily by voice commands and can be used to obtain the desired result in a quicker span of time and much easily.

### C. In hospitals:

This assistant can be used extensively in the hospitals where it can be used in surgical operations. Robotic arm has been used in various surgeries across hospitals [8]. Furthermore, if it can be guided by the voice commands and carry out the specified task, efficiency can be increased thus also causing the human labour to reduce.

## VIII. SUMMARY AND CONCLUSION

Voice control for a home assistant robot is developed in this paper. The voice commands are processed in real-time, using an offline server. The speech signal commands are directly communicated to the server over a wired network. The personal assistant robot is developed on a micro-controller based platform and can be aware of its current location. Performance evaluation is carried out with encouraging results of the initial experiments. Possible improvements are also discussed towards potential applications in home, hospitals, car systems and industries.

The effect of the distance between the mouth and microphone on the robot, the performance of the robot, effect of noise on the speech to text conversion are some of the areas that can be further explored. The accent of the speaker does not affect the operation of the robot as the voice commands are processed using a cloud server which functions irrespective of the accent of the speaker.

# Robot Voice A Voice Controlled Robot using Arduino

Using renewable source of energy for the functioning of the robot would not only improve upon the cost of the robot but would also prove to be eco-friendly. Solar cells can be a possible source of energy that can be used. The robotic assistant developed has potential applications ranging from chemical industries to comfortable scenario inside homes. This paper should be helpful in showcasing a server based application in developing a voice-controlled robotic assistant.

## REFERENCES

1. A Voice-Controlled Personal Robot AssistantAnurag Mishra, Pooja Makula, Akshay Kumar, Krit Karan and V.K. Mittal, IIIT, Chittoor, A.P., India.
2. H.Uehara,H. HigaandT.Soken,"A Mobile Robotic Armfor people with severe disabilities", International Conference on Biomedical Robotics and Biomechatronics (BioRob), 3rd IEEE RAS and EMBS , Tokyo, pp. 126- 129, September 2010, ISSN:2155-1774.
3. David Orenstein, "People with paralysis control robotic arms using brain", https://news.brown.edu/articles/2012/05/braingate2 (Last viewed on October 23, 2014).
4. Lin. H. C, Lee. S. T, Wu. C. T, Lee. W. Y and Lin. C. C, "Robotic Arm drilling surgical navigation system", International conference on Advanced Robotics and Intelligent Systems (ARIS), Taipei, pp. 144-147, June 2014.
5. Rong-Jyue Wang, Jun-Wei Zhang, Jia-Ming Xu and Hsin-Yu Liu, "The Multiple-function Intelligent Robotic Arms", IEEE International Confer- ence on Fuzzy Systems, FUZZ-IEEE, Jeju Island, pp. 1995 - 2000, August 2009, ISSN:1098-7584.
6. "Programming Arduino Getting Started with Sketches". McGraw-Hill. Nov 8, 2011. Retrieved 2013-03-28.
7. http://www.bitsophia.com/en-US/BitVoicerServer/Overview.aspx.
8. The National Staff, "Robot arm performs heart surgeries at Sharjah hospital", http://www.thenational.ae/uae/health/robot-arm-performs-heart- surgeries-at-sharjah-hospital (Last viewed on November 13, 2014).

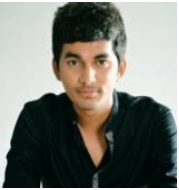
**Mr. Vineeth Teeda**, was completed his B.Tech degree in 2016 at NRI Institute of Technology, Department of Electronics and Communication Engineering.

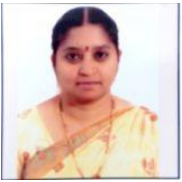
**Mrs. K. Sujatha,** working as Associate Professor in Electronics and Communication Engineering, in NRI Institute of Technology, Guntur, A.P, India.

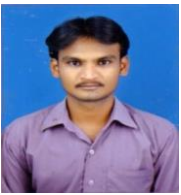
**Mr. Rakesh Mutukuru** working as Assistant Professor in Electronics and Communication Engineering, in NRI Institute of Technology, Guntur, A.P, India.